\documentclass[conference]{IEEEtran}
\usepackage[pdftex]{graphicx}

\usepackage{amsmath}
\ifCLASSINFOpdf
\else
\fi
\begin{document}
%
\title{Online Vehicle Detection For Estimating Traffic Status }

\author{\IEEEauthorblockN{Ranch Y.Q. Lai}
\IEEEauthorblockA{Department of Computer Science\\
Hong Kong Baptist University\\
Email: yqlai@comp.hkbu.edu.hk}
}


%


\maketitle

\begin{abstract}
We propose a traffic congestion estimation system based on unsupervised on-line learning algorithm. The system does not rely on background extraction or motion detection. It extracts local features inside detection regions of variable size which are drawn on lanes in advance. The extracted features are then clustered into two classes using K-means and Gaussian Mixture Models(GMM). A Bayes classifier is used to detect vehicles according to the previous cluster information which keeps updated whenever system is running by on-line EM algorithm.  Experimental result shows that our system can be adapted to various traffic scenes for estimating traffic status.
\end{abstract}


%
\IEEEpeerreviewmaketitle

\section{Introduction}

As vision-based traffic monitoring systems become more and more prevalent due to their low-cost and easy-to-deploy aspects, research on application of computer vision to traffic state measurement  attracts more interest than before. Although the history of traffic monitoring system can be at least traced back to 1991\cite{michalopoulos1991vehicle} and new approaches are emerging\cite{bragatto2008new}, various problems are still to be tackled. Situation is much worse in urban areas, where congestion most likely occurs, especially at the intersections of downtowns. Monitoring the urban scenes and highways has been extensively studied. Most of the approaches are based on background extraction (or motion detection) and vehicle tracking\cite{beymer1997real}\cite{hsu2005real}\cite{morris2007real}\cite{morris2008learning}.  These approaches are common and  prevalent because 1) background can be easily extracted because of light traffic condition. 2) Tracking is straightforward as targets are readily segmented. 3)  Trajectories will be available for even analysis such as retrogradation, aberrance, clash, etc.  

However, research on highways and suburban areas cannot necessarily apply to urban areas, in which there are two main problems need to be solved if we rely on vehicle tracking. One is that background rarely reveals in most situations of urban environment. For example, at the intersections, the whole lanes may be dominated by vehicles in waiting queue for a very long time as vehicles go and come. The other reason is that, tracking is difficult (if not impossible) because of the difficulty in vehicle segmentation when traffic jam happens, even though the background information may be still pure. In this case, vehicles are so heavily (partial or total) occluded that most of the tracking strategies fail to work because of difficulty in segmentation\cite{jung1999traffic}. Blob tracker\cite{song2005model} is likely to link two or more vehicles together either by region or by silhouette\cite{song2005model}. Point tracker\cite{beymer1997real}\cite{yang2007tracking} may falsely group features from different targets. Instead of taking these approaches which rely on background and segmentation, we propose a novel approach to measure traffic state at the congested intersections.
 
 \begin{figure}
  \includegraphics[width=\linewidth]{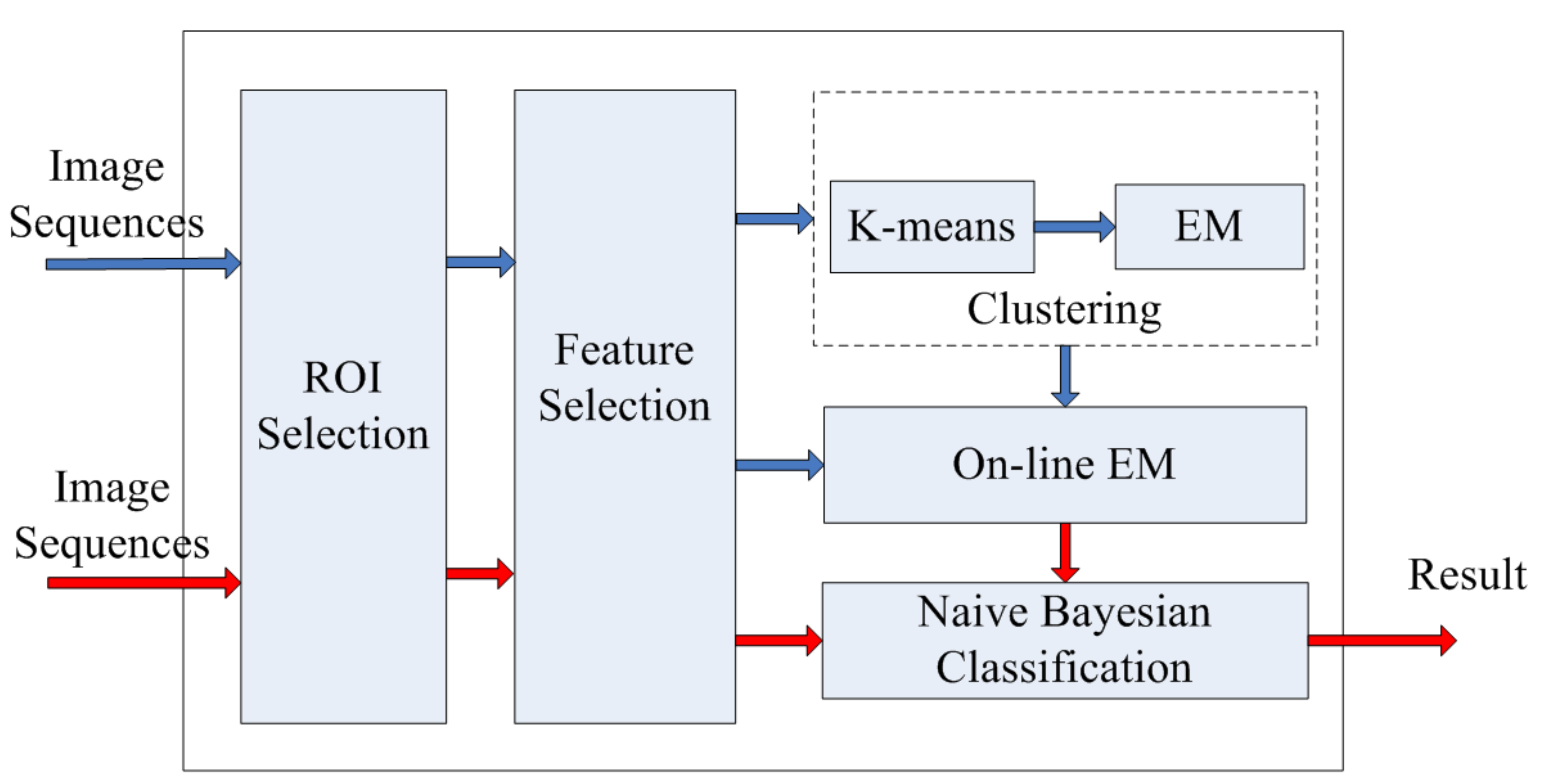}
  \caption{Block diagram of our traffic state monitoring system.The blue arrows show the learning path whereas the red arrows show the classification path}
  \label{framework}
 \end{figure}
 
Our system does not involve background extraction and vehicle segmentation. Instead, we classify vehicles from lanes directly through an online learning scheme and Bayes classifier. The block diagram of our system is showed in figure \ref{framework}.   In the ROI selection phase, lanes are divided into blocks. Features are extracted from these blocks and fed into a learning model, which consists of an unsupervised clustering phase in the initialization stage and an on-line expectation-maximum (EM) learning phase that runs after the initialization.  Considering that the initialization stage should be short in duration and the on-line learning rate is relatively much lower, we  do not merge them into one, although it is technically possible to do that.  

The rest of this paper is organized as follows: Section II is a brief introduction to the related work. Section III discusses ROI section. Section IV focuses on Feature selection. In Section V we will talk about Clustering and On-line learning. Section VI is Bayesian classification. The experimental result will given in section VII followed by a conclusion section.

\section{RELATED WORK}
Zanin\cite{zanin2003efficient} proposed a vehicle queue detection scheme based on background subtraction and movement analysis. The need for background information made it difficult to handle usually congested scenes like intersection.  Porikli\cite{porikli2004traffic} Proposed traffic congestion estimation scheme using Gaussian Mixture Hidden Markov models. They extracted features from MPEG compressed domain and trained a set of GM-HMM to estimate traffic state. Kato\cite{kato2002hmm} used HMM-based segmentation method to classify objects in traffic scenes as shadows, foreground or background objects.
Our approach do not rely on background extraction and motion analysis, which means that it can work well even if the vehicle queue is completely stop for a long duration. Another difference is that we divide lanes into blocks, and detect vehicles inside the blocks. This will eliminate the necessity for vehicle segmentation and the measurement for the length of vehicle queue is as easy as counting the blocks that contain vehicles. 

\section{ROI SELECTION}
\begin{figure}
\centering
\includegraphics[width=0.4\linewidth]{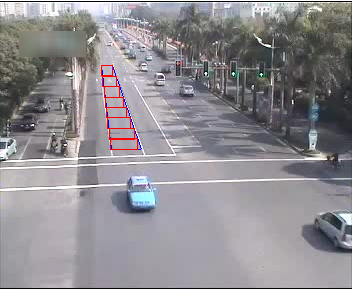}
\includegraphics[width=0.4\linewidth]{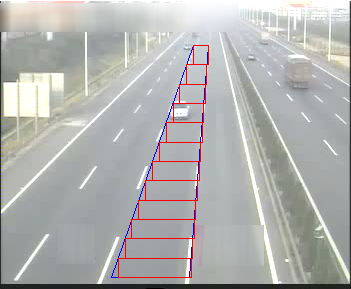}
\caption{ROI demonstration.  Rectangles of Various size are used to approximate trapezoidal lane. Each Rectangle is treated as an independent sub-image in later processing.}
\label{roi}
\end{figure}

The ROI selection scheme is pretty simple and straightforward here. As the length of the waiting queue is the most important parameter at the intersection, we need to detect the existence of vehicles both near the pavement and at the far end. Due to the position of traffic cameras and perspective projection, the far end of lanes is narrower than the front end. Therefore, we draw the outline of lanes which is similar to trapezoid. We then divide the trapezoid into several rectangles; each of these rectangles is called Region of interest (ROI). All the features are extracted within these ROI. See figure \ref{roi} for an example.

\section{FEATURE SELECTION}
For a classier to work in really applications, multiple features are usually needed. AS the complexity of the classifiers depends on the dimension of the feature vector, we shall select those features that are most discriminant.  What's more, the feature selection algorithms should be as simple as possible for maximally reducing the system overall complexity.  During our experiments, we found that basic image features including local Entropy, edge and moment are best choice for our system to work in real-time.  Since we are using Naive Bayes classifier,  the ease of normalization of each feature has been taken into account when we choose these features.

\subsection{Maximum Local Entropy}
Entropy is a good measurement for texture and common used in object detection\cite{hsu2005real} \cite{shi2009robust}. It is defined as $E =  - \sum\limits_{i = 1}^L {p(i){{\log }_2}p(i)} $, where $\sum\limits_{i = 1}^L {p(i)}  = 1,p(i) >  = 0,i = 1,...,L$. The probability is naturally  represented by histogram, i.e., $p(i) = \frac{1}{{wh}}\sum\limits_{x = 1}^w {\sum\limits_{y = 1}^h {1\{ I(x,y) = i\} } } $ where$w$  and $h$ are width and height of the image patch respectively. and $1\{ \cdot \} $ is the indicator function.

\begin{figure}[!t]
\centering
\includegraphics[width=\linewidth]{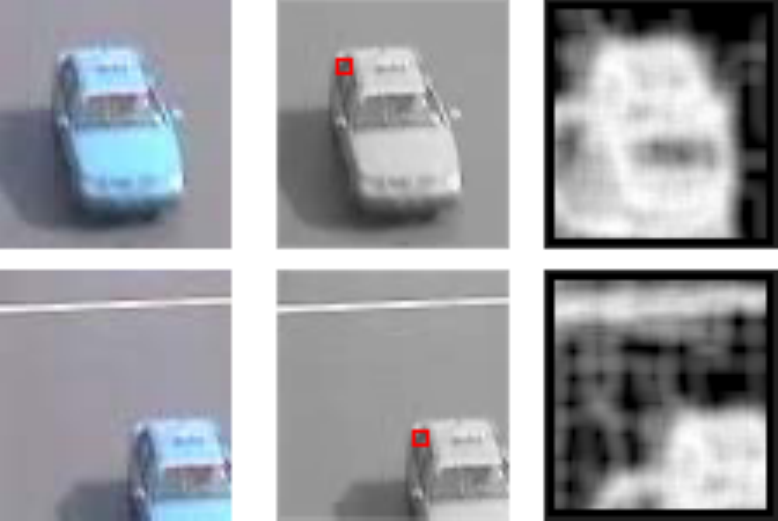}

\caption{Local entropy measurement of two images. 
\label{entropy}
The first column shows two different images containing
the same vehicle fully and partially. The second 
column shows the location of maximum local entropy 
measurement, denoted by two red squares. Note that 
the two red squares are in the same position relative to 
the vehicle.  The last column shows local measurement 
of the two images as gray scale images}
\end{figure}

It is easy to find that the maximum entropy is achieved to ${\log _2}(L)$ if and only if $p(i) = \frac{1}{L}$ for $i = 1,...,L$, which means that scattered gray-levels lead to large entropy while a clean background has a zero entropy. It is therefore discriminative to use entropy as a feature,
 	
While entropy is a good measure of texture, applying entropy directly to a whole image will obscure its details. Two different images may have the same entropy value, while two images both contain the same object may have two different entropy values. For example, in figure \ref{entropy}, images of the first column both contain the same vehicle (although the first image has a clear view and the second one has only a partial view). The entropy of the first image is 5.5953 while the entropy of the second one is 	6.2018. In order to best describe the similarity of these two images, we induce a local entropy measurement,
\begin{equation}
E' = \max \limits_{x,y} \{ { - \sum\limits_{i = 1}^L {p(i,x,y){{\log }_2}p(i,x,y)} }\} 
\end{equation}
where $x = 1,...,m$,$y = 1,...,n$ and
\begin{equation}
p(i,x,y) = \frac{1}{{(2w' + 1)(2h' + 1)}}\sum\limits_{j = x - w'}^{x + w'} {\sum\limits_{k = y - h'}^{y + h'} {1\{ I(j,k) = i\} } } 
\end{equation}

$2w' + 1$ and $2h' + 1$ are the window width and height respectively. $w'$ and $h'$  vary from 1 to 3 in different scenes and resolutions.The local entropy measurement is illustrated in Figure \ref{entropy}.  

\subsection{Edge }

Unlike those tackle complete vehicle detection, we take edge as another measurement of texture here.  Four different types of edge detection kernel are defined as follows

${h_1} = \left[ {\begin{array}{*{20}{c}}
  { - 1}&{ - 2}&{ - 1} \\ 
  0&0&0 \\ 
  1&2&1 
\end{array}} \right]$,
${h_2} = \left[ {\begin{array}{*{20}{c}}
  { - 1}&0&1 \\ 
  { - 2}&0&2 \\ 
  { - 1}&0&1 
\end{array}} \right]$,

${h_3} = \left[ {\begin{array}{*{20}{c}}
  0&{ - 1}&{ - 1} \\ 
  1&0&{ - 1} \\ 
  1&1&0 
\end{array}} \right]$,
${h_4} = \left[ {\begin{array}{*{20}{c}}
  { - 1}&{ - 1}&0 \\ 
  { - 1}&0&1 \\ 
  0&1&1 
\end{array}} \right]$

${h_1}$ and ${h_2}$ are  the well-known Sobel detectors, which detect horizontal edges respectively; ${h_3}$  and ${h_4}$  detect left diagonal edges and right diagonal edges respectively.   After convolving the above kernels with a gray-scale image, we obtain an edge image ${I_{{h_t}}}$. The edge image is further processed by the operation : 
\begin{equation}
\Gamma ({h_t}) = \frac{1}{{wh}}\sum\limits_{x = 1}^w {\sum\limits_{y=1}^h {1\left\{ {{I_{{h_t}}}(x,y) > \varphi } \right\}} } 
\end{equation}

\subsection{Other measurements}
We further define three measurements. The first one is rate of non-zero histogram bins, or $B$, which is defined as
$B = \frac{1}{L}\sum\limits_{i = 1}^{L} {1\{ {p(i) > 0} \}} $. 
 Another feature is normalized first moment, which is defined as 
${M_1} = \frac{1}{L}\sum\limits_{i = 1}^L {i \cdot p(i)} $. ${M_1}$is simply the mean of image intensity.  The last feature is the second moment of image:
\begin{equation}
{M_2} = \sum\limits_{i = 1}^L {\left( {{i^2}p(i)} \right)}  - {M_1}^2
\end{equation}
To normalize ${M_2}$, we need to find its maximum value. Another definition of ${M_2}$  is ${M_2} = \frac{1}{p}\sum\limits_{i = 1}^n {x_i^2}  - {(\frac{1}{p}\sum\limits_{i = 1}^n {{x_i}} )^2} = {{\mathbf{x}}^t}(\frac{I}{p} - \frac{{{\mathbf{b}}{{\mathbf{b}}^t}}}{{{p^2}}}){\mathbf{x}}$
where ${\mathbf{b}} = {[1,...,1]^t}$ . We treat an image as a $n \times 1$ vector  and ${\mathbf{x}}$ is ${x_i}$ its $i{\text{th}}$  component. To find the maximum value of  (11) is a quadratic programming subjected to $1 \leq{x_i} \leq L$   for  $i = 1,...n$. Without further derivation, we give the maximum value here directly, that is$\frac{{{{(L - 1)}^2}}}{4}$ .Therefore, normalized version of equation (10) is  
${M_2} = \frac{4}{{{{(L - 1)}^2}}}\left( {\sum\limits_{i = 1}^L {\left( {{i^2}p(i)} \right)}  - {M_1}^2} \right)$
\subsection{Selection criterion}
\begin{figure}[!t]
\centering
\includegraphics[width=\linewidth]{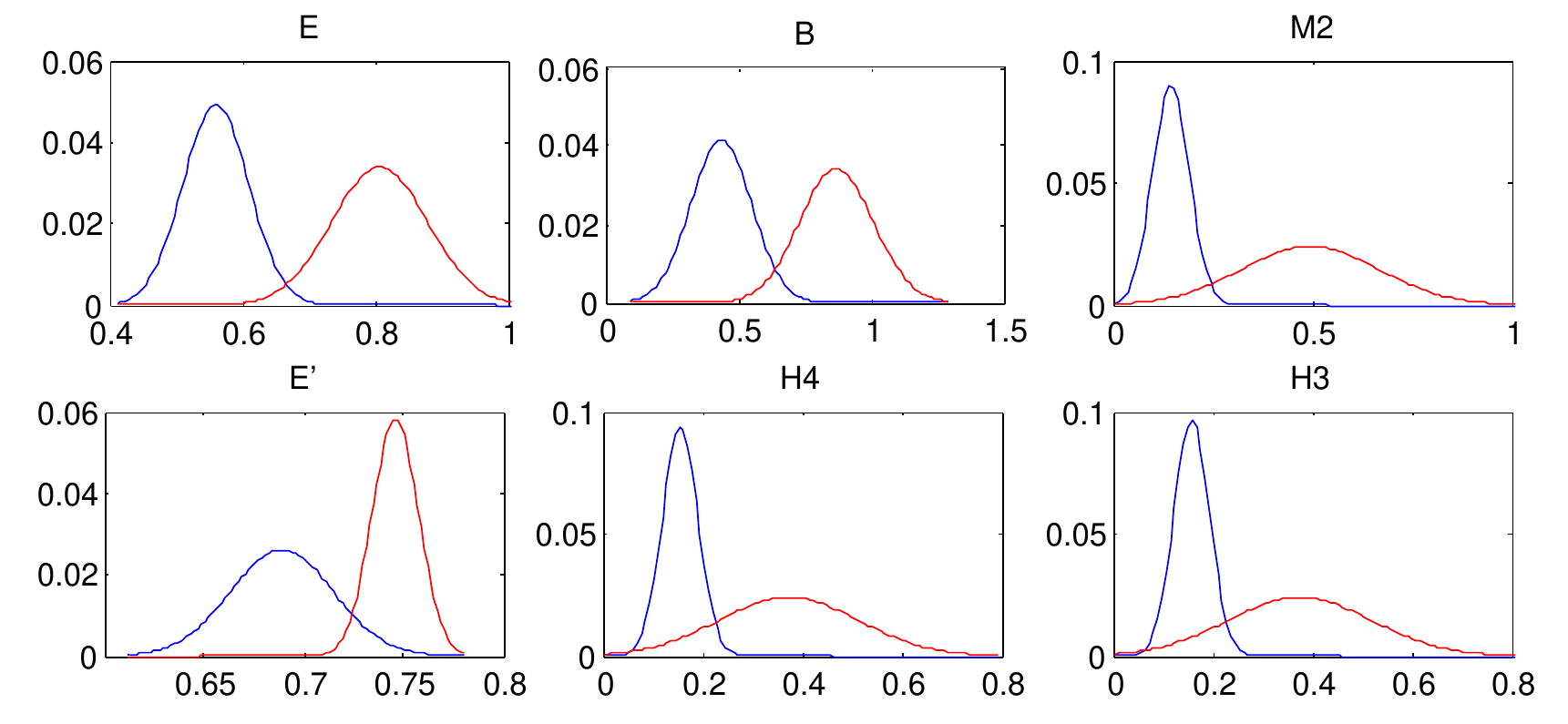}

\caption{The red curve and blue curve represent the distribution of each feature extracted from training samples which contain vehicles and those contain only background, respectively. These visualized distributions provide an intuition on which features are superior for classification. The title of each chart follows symbols given in table I.}
\label{featuredistribution}
\end{figure}

We test each of these measurement using two sets of samples which are manually labelled beforehand. The quality of each feature is judged by Fisher’s criterion,  $J = \frac{{{{({\mu _1} - {\mu _2})}^2}}}{{\sigma _1^2 + \sigma _2^2}}$
where${\mu _i}$ and ${\sigma _i}^2$ are mean and variance of feature value of class  respectively. Large   means large inter-class distance and  small intra-class variation. All the feature measurements are tested on a training set containing 4000 vehicle images and 2000 lane images according to Fisher’s criterion. The testing result is given in table 1, and Fig \ref{featuredistribution} shows the separating distance in the form of probability density function (PDF).

\begin{table}[!t]

\caption{An Example of a Table}

\label{table_example}
\centering
\begin{tabular}{|c|c|c|}
\hline
Symbol & Meaning & Separating distance\\
\hline\hline
$E$ & Entropy & 2.0267\\
\hline
$B$ & Total non-zero histogram bins & 1.7142\\
\hline
${M_2}$ & Second order central moment & 1.5166\\
\hline
$\Gamma ({h_4})$ & Right diagonal edge &1.2329\\
\hline
$\Gamma ({h_3})$&Left diagonal edge&1.2047\\
\hline
${M_1}$&First order moment(mean)&0.8501\\
\hline
$\Gamma ({h_2})$&Vertical edge&0.7979\\
\hline
$\Gamma ({h_1})$&Horizontal edge&0.8353\\
\hline
\end{tabular}
\end{table}

Note that all features are normalized before being assembled as a feature vector. As we use Euclidean distance, non-normalized feature vector will lead to focus on larger scale feature component.

\section{CLUSTERING AND ONLINE EM ALGORITHM}
\subsection{Clustering}
We use the standard K-means clustering algorithm \cite{michalopoulos1991vehicle} to coarsely estimate the cluster centroids which serves as an initial guess for the subsequent EM Clustering algorithm. Without discussing algorithm details, we give the K-means parameter values in our case here directly.
1.	K=2, i.e. there are two clusters: vehicle and lane
2.	Cluster distance is measured by Euclidean distance
3.	Repeat the algorithm N times to avoid trapping into local maximums.(N=3 is sufficient)

\subsection{EM for GMM }
The output of K-means algorithm are two cluster centroids, as well as labels of data indicating which cluster they belong too. The two clusters are further modelled with Gaussian distribution, i.e.
$p\left( {x|x \in c;{\mu _c},{\Sigma _c}} \right) = \frac{1}{{{{\left( {2\pi } \right)}^{\frac{n}{2}}}|{\Sigma _c}|}}\exp \left( { - \frac{1}{2}{{(x - {\mu _c})}^T}\Sigma _c^{ - 1}(x - {\mu _c})} \right)$

where $c \in \left\{ {l,v} \right\}$  represents  lane and vehicle respectively. $x \in {\mathbf{R}^n}$  is the feature vector.  ${\mu _c} \in {\mathbf{R}^n}$ is the mean vector, and ${\Sigma _c} \in S_{ +  + }^n$ denotes the covariance matrix, which is positive definite. We further simplify the covariance matrix as a diagonal matrix to lower computation complexity without much loss of accuracy. 

     Suppose that we have $m$ feature vectors which we want to divide into two classes and model them with two separate Gaussian distributions. To put it another way, the problem is to find  ${\mu _c}$,  ${\Sigma _c}$ and more accurately determine memberships of all data to the two models . By using the well-known EM algorithm, we can easily solve this mixture models problem [5]. First, we define the class prior density as 
     $p(x \in c) = {\Phi _c}$
     with $\sum\nolimits_{c = \{ l,v\} } {{\Phi _c}}  = 1$ and ${\Phi _c} > 0$.

1.	E-step:
For every $i$  and $c$ , estimate memberships according to  
\begin{eqnarray}
\omega _c^{^{(i)}} &=& p( {{x^{(i)}} \in c|{x^{(i)}};{\Phi _c},{\mu _c},{\Sigma _c}} ) \nonumber\\
&=& \frac{1}{D}	 p( {{x^{(i)}}|{x^{(i)}} \in c;{\mu _c},{\Sigma _c}} )p({x^{(i)}} \in c;{\Phi _c})\label{omega}
\end{eqnarray}

2.	M-step:
\begin{eqnarray}
&&{\Phi _c} = \frac{1}{m}\sum\limits_{i = 1}^m {\omega _c^{(i)}} \\
&&{M_c} = \sum\limits_{i = 1}^m {\omega _c^{(i)}} \\
&&{\mu _c} = \frac{1}{{{M_c}}}\sum\limits_{i = 1}^m {\omega _c^{(i)}{x^{(i)}}} \\
&&{\Sigma _c} = \frac{1}{{{M_c}}}\sum\limits_{i = 1}^m {\omega _c^{(i)}({x^{(i)}} - {\mu _c})} {({x^{(i)}} - {\mu _c})^T}
\end{eqnarray}

	Then, a common used EM-based GMM algorithm is listed here without proving it. 
	
\begin{figure}[!t]
\centering
\includegraphics[width=0.4\linewidth]{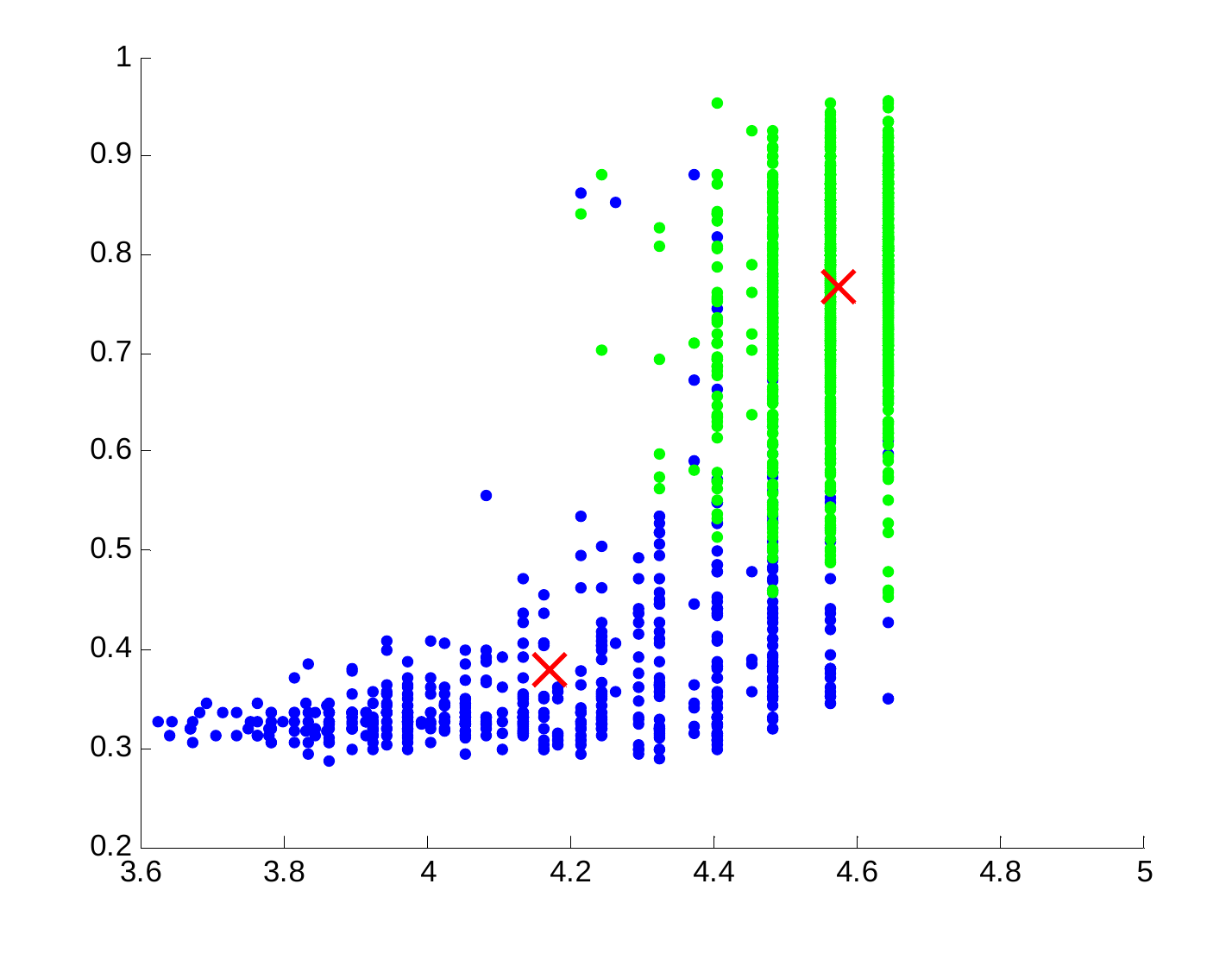}
\includegraphics[width=0.4\linewidth]{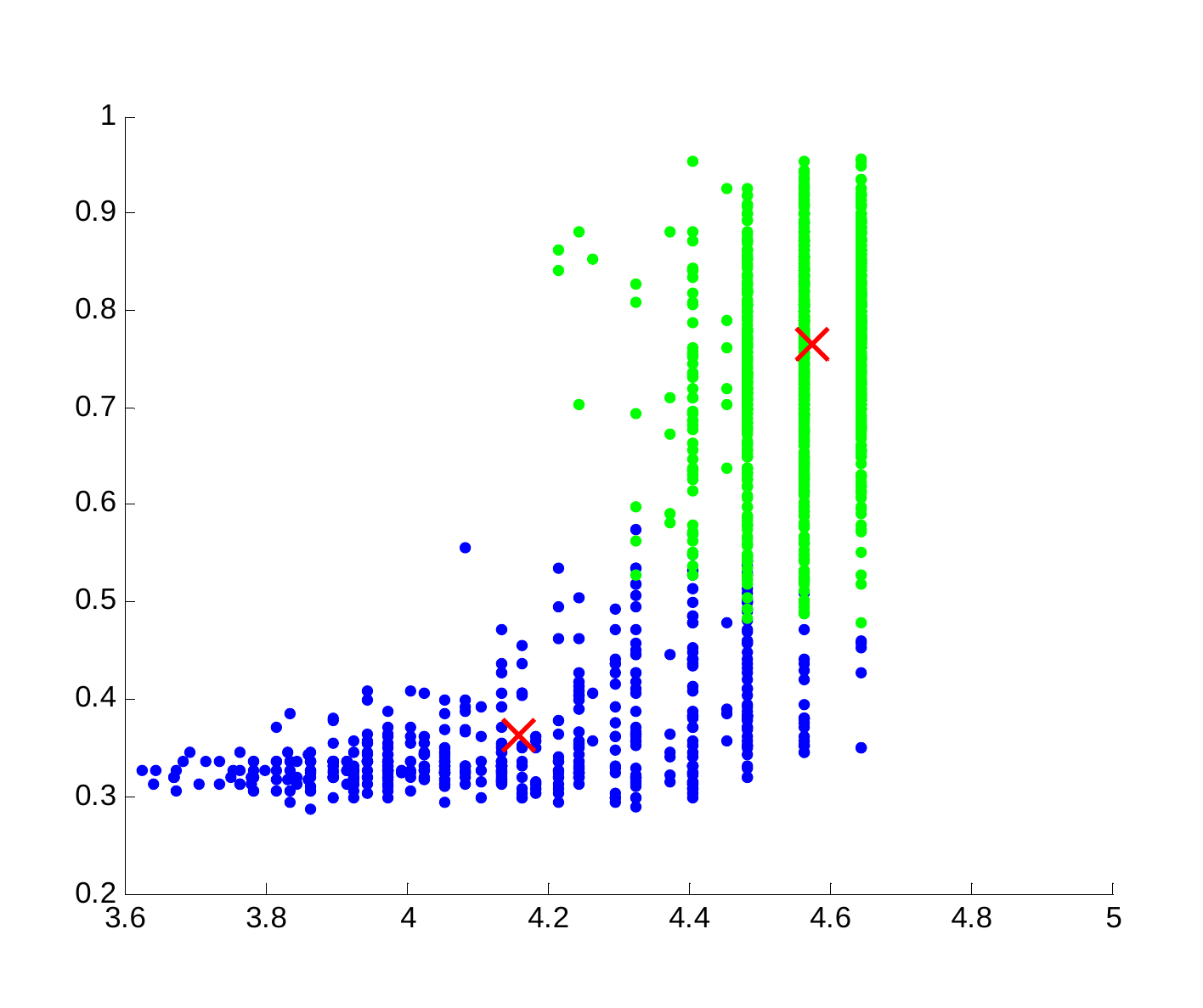}

\caption{Clustering result for  the first two dimensions of feature vectors are demonstrated. The left one is k-means clustering result; The right one is EM clustering result. It can be seen that there are considerable difference between them. }
\label{clustering}
\end{figure}

	In the above algorithm,  ${x^{(i)}}$ denotes the  $i{\text{th}}$ feature vector. $D$  is denominator which normalizes $\omega _c^{^{(i)}}$  so  that $\sum\nolimits_c {\omega _c^{^{(i)}}}  = 1$  for every $i = 1,...,m$ .The initial value for ${\Phi _c}$  is 0.5, for  ${\mu _c}$ is the output of cluster centroids from k-means, and for ${\Sigma _c}$ is just the identity matrix. A snapshot of k-means and EM clustering algorithm is taken from our system and showed in figure \ref{clustering}, which shows the cluster distribution of the first two components in the feature space.
	
\section{	ON-LINE EM}
Suppose that we already have  samples in hand, and the GMM is well modeled accordingly. As the monitoring system runs, the environment (mostly the lighting condition) will gradually change. If no updating is made, the system probably will fail when the change becomes obvious. Therefore, an on-line learning phase is adopted to update the parameters of GMM. 

Derivation of the on-line EM algorithm is intuitive. Let  ${x^{(m + 1)}}$ denotes the incoming sample, following the notations of previous EM algorithm, we have the following on-line learning algorithm,

1.	Estimate  $\omega _c^{^{(m + 1)}}$ according to \eqref{omega}.

2.	Update model parameters 
\begin{eqnarray}
&&{\Phi _{c,m + 1}} = \frac{1}{{m + 1}}( {{\Phi _{c,m}}m + \omega _c^{(m + 1)}} )\\
&&{M_{c,m + 1}} = {M_{c,m}} + \omega _c^{(m + 1)}\\
&&{\mu _{c,m + 1}} = \frac{1}{{{M_{c,m + 1}}}}( {{\mu _{c,m}}{M_{c,m}} + \omega _c^{(m + 1)}{x^{(m + 1)}}} )\\
&&{\Sigma _{c,m + 1}} = \frac{1}{M_{c,m + 1}}( \Sigma _{c,m}M_{c,m} + \nonumber \\
&& \omega _c^{(m + 1)}(x^{(m + 1)} - \mu _{c,m + 1})(x^{(m + 1)} - \mu _{c,m + 1})^T )
\end{eqnarray}

	One problem of On-line EM is that, when m is too large, ${M_{c,m}}$ will be also too large that the incoming new sample barely influences the model parameters. To avoid this, we fix  $\omega _c^{(m + 1)}/{M_{c,m}} = \lambda  = m/(m + 1)$ and then adjust the model updating equations accordingly.
	
\section{BAYES CLASSIFIER}
Given a sample feature vector   and two class models that have been trained, according to the Bayes rule, 
\begin{equation}
p(x \in c|x;{\theta _c}) = \frac{{p(x|x \in c;{\theta _c})p(x \in c;{\theta _c})}}{{\sum\nolimits_c {p(x|x \in c;{\theta _c})p(x \in c;{\theta _c})} }}\label{bayes}
\end{equation}
for $c \in \{ l,v\} $.$p(x|x \in c;{\theta _c}) = N(x;{\mu _c},{\Sigma _c})$is the likelihood term. ${\theta _c}= [{\mu _c},{\Sigma _c},{\Phi _c}]$  . $p(x \in c;{\theta _c}) = {\Phi _c}$ is the prior density. A feature vector  is said to be from class $v$  if $p(x \in v|x;{\theta _v}) > p(x \in l|x;{\theta _l})$.

Another form of \eqref{bayes} is the discriminate function 		
$f(x) = \log p(x \in v|x;{\theta _v}) - \log p(x \in l|x;{\theta _l}) > 0$ which is further simplified as 
\begin{eqnarray}
f(x) &=& \log \frac{{|{\Sigma _l}|}}{{|{\Sigma _v}|}} + \log \frac{{{\Phi _v}}}{{{\Phi _l}}} + \frac{1}{2}{(x - {\mu _l})^T}\Sigma _l^{ - 1}(x - {\mu _l})\nonumber\\
&& - \frac{1}{2}{(x - {\mu _v})^T}\Sigma _v^{ - 1}(x - {\mu _v})\label{logdiscr}
\end{eqnarray}

Every incoming feature is classified to be vehicle or lane according to \eqref{logdiscr}.

\section{EXPERIMENTAL RESULT}
Although our video sources contain hue information, we utilized only the intensity  channel. The image sequence are  in size and 25FPS. Because high chronological preciseness is not necessary, the video rate is down-sampled to 5FPS, which lowers the consumption of computational resource tremendously.   
	The values of aforementioned parameters are sum up in table \ref{table2}.
	
\begin{table}[!t]
\caption{SUMMARY OF PARAMETER VALUES}
\label{table2}
\centering
\begin{tabular}{|c|c|c|c|}
\hline
parameter & value & parameter & value\\
\hline\hline
$L$ & 256 & $w'$ & 2 \\ 
\hline $\varphi $  & 30 & $h'$ & 2 \\ 
\hline $w$ & 352 & $L'$ & 32 \\ 
\hline $h$ & 288 & $\lambda $ & 0.05 \\ 
\hline
\end{tabular}
\end{table}

\section{Conclusion}
In this paper, we have proposed a congestion estimation system which has wide application in urban traffic scenes. By using a traffic monitoring camera, it can effectively estimate the current traffic state, measure the length of vehicle queue and provide useful information for the traffic control departments. 
We simplify the flow chart of our system by dividing lanes into blocks, which eliminates the need for vehicle segmentation. However, it brings up another new challenge, that is, to detect partly visible objects. By incorporating GMM and Bayesian into our system, we have successfully tackled the detection of vehicle existence, assuming that only lanes and vehicles are presented. In the future research,  vehicle type recognition and pedestrian  detection within lane blocks will be of great interest to us.

\begin{figure}[!t]
\centering
\includegraphics[width=0.3\linewidth]{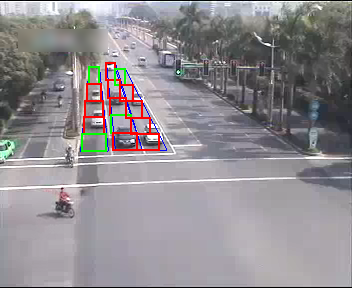}
\includegraphics[width=0.3\linewidth]{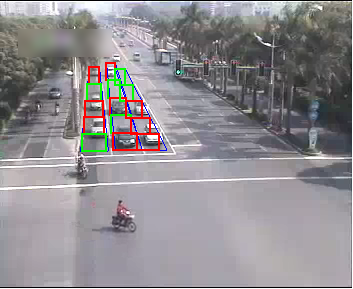}
\includegraphics[width=0.3\linewidth]{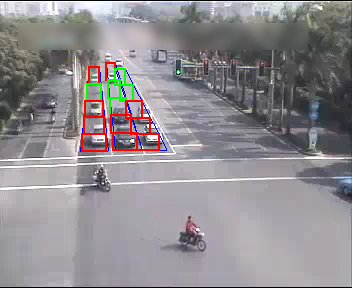}

\includegraphics[width=0.3\linewidth]{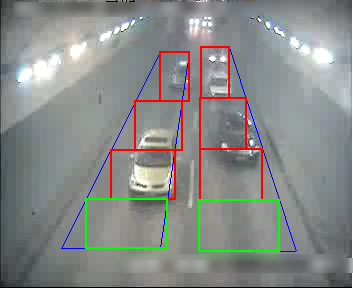}
\includegraphics[width=0.3\linewidth]{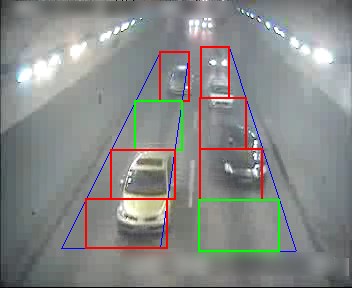}
\includegraphics[width=0.3\linewidth]{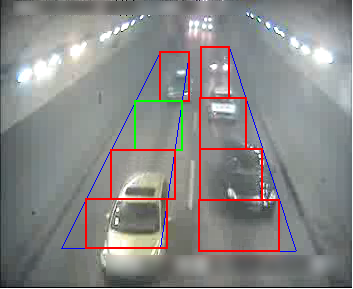}

\includegraphics[width=0.3\linewidth]{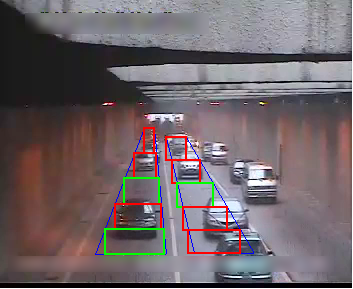}
\includegraphics[width=0.3\linewidth]{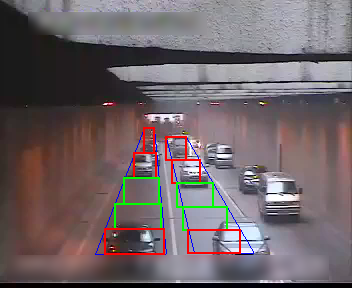}
\includegraphics[width=0.3\linewidth]{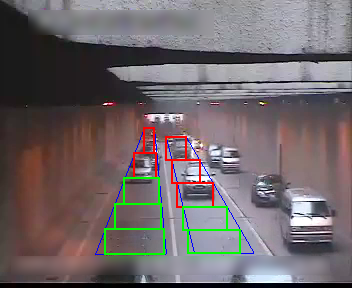}

\includegraphics[width=0.3\linewidth]{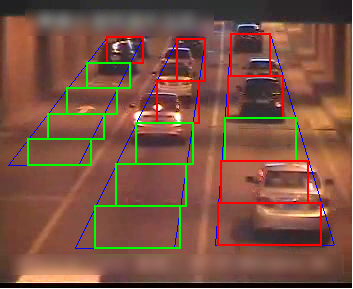}
\includegraphics[width=0.3\linewidth]{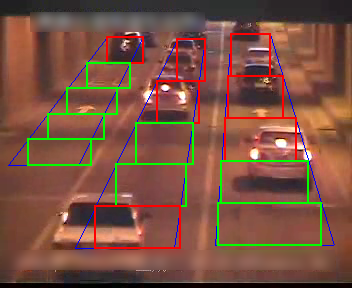}
\includegraphics[width=0.3\linewidth]{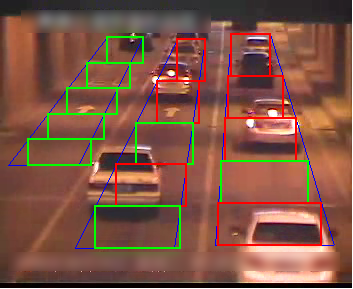}

\caption{Four different scenes are presented here. The red square means vehicle exists whereas the green square means otherwise. }
\label{experimentresult}
\end{figure}


\section*{Acknowledgment}

\bibliographystyle{IEEEtran}
\bibliography{reference}

\end{document}